\documentclass[10pt,letterpaper]{article}

\usepackage[pagenumbers]{cvpr} 

\usepackage{graphicx}

\usepackage{amsmath}
\usepackage{float}
\usepackage{tabularx}
\usepackage{multicol}
\usepackage{listings}
\usepackage{xcolor}
\usepackage{placeins}
\usepackage{enumitem}

\newcommand{\sygra}{\textsc{SyGra}\xspace}

\usepackage[T1]{fontenc}

\usepackage[T1]{fontenc}
\usepackage[utf8]{inputenc}
\usepackage{multirow}
\usepackage{booktabs}
\usepackage{pifont}
\usepackage{threeparttable}
\usepackage[numbers,sort,compress]{natbib}

\usepackage[pagebackref,breaklinks,colorlinks]{hyperref}
\usepackage[capitalize]{cleveref}
\crefname{section}{Sec.}{Secs.}
\Crefname{section}{Section}{Sections}
\Crefname{table}{Table}{Tables}
\crefname{table}{Tab.}{Tabs.}
\newcommand{\xmark}{\textcolor{red}{\ding{55}}}
\newcommand{\checkmark}{\textcolor{green}{\ding{51}}}

\definecolor{delim}{RGB}{20,105,176}
\definecolor{numb}{RGB}{106, 109, 32}
\definecolor{string}{rgb}{0.64,0.08,0.08}

\lstdefinelanguage{JSON}{
    frame=single,
    showspaces=false,
    showstringspaces=false,
    showtabs=false,
    breaklines=true,
    postbreak=\raisebox{0ex}[0ex][0ex]{\ensuremath{\color{gray}\hookrightarrow\space}},
    breakatwhitespace=true,
    basicstyle=\ttfamily\small,
    upquote=true,
    morestring=[b]",
    stringstyle=\color{string},
    literate=
     *{0}{{{\color{numb}0}}}{1}
      {1}{{{\color{numb}1}}}{1}
      {2}{{{\color{numb}2}}}{1}
      {3}{{{\color{numb}3}}}{1}
      {4}{{{\color{numb}4}}}{1}
      {5}{{{\color{numb}5}}}{1}
      {6}{{{\color{numb}6}}}{1}
      {7}{{{\color{numb}7}}}{1}
      {8}{{{\color{numb}8}}}{1}
      {9}{{{\color{numb}9}}}{1}
      {\{}{{{\color{delim}{\{}}}}{1}
      {\}}{{{\color{delim}{\}}}}}{1}
      {[}{{{\color{delim}{[}}}}{1}
      {]}{{{\color{delim}{]}}}}{1},
}

\lstdefinestyle{yaml}{
     basicstyle=\color{teal}\footnotesize,
     rulecolor=\color{black},
     string=[s]{'}{'},
     stringstyle=\color{red},
     comment=[l]{:},
     commentstyle=\color{black},
     morecomment=[l]{-}
 }

\def\confName{NOWAI}
\def\confYear{2025}
\def\cvprPaperID{NOWAI071} 

\providecommand{\keywords}[1]{\textbf{\textit{Keywords: }} #1}
\newlength\savewidth

\begin{document}

\title{\sygra: A Unified Graph-Based Framework for Scalable Generation, \\ Quality Tagging, and Management of Synthetic Data}

\author{
 \textbf{Bidyapati Pradhan} \quad \textbf{Surajit Dasgupta} \quad \textbf{Amit Kumar Saha} \quad \textbf{Omkar Anustoop}\\
 \textbf{Sriram Puttagunta} \quad \textbf{Vipul Mittal} \quad \textbf{Gopal Sarda}\\
  \\ 
 ServiceNow Inc.\\
  \tt\small \{bidyapati.pradhan, surajit.dasgupta, amit.saha, omkar.anustoop, \\
  \tt\small sriram.puttagunta, vipul.mittal, gopal.sarda\}@servicenow.com
}

\maketitle

\begin{abstract}
The advancement of large language models (LLMs) is critically dependent on the availability of high-quality datasets for Supervised Fine-Tuning (SFT), alignment tasks like Direct Preference Optimization (DPO), etc. In this work, we present a comprehensive synthetic data generation framework, \sygra -- that facilitates scalable, configurable, and high-fidelity generation of synthetic data tailored for these training paradigms. Our approach employs a modular and configuration-based pipeline capable of modeling complex dialogue flows with minimal manual intervention. This framework uses a dual-stage quality tagging mechanism, combining heuristic rules and LLM-based evaluations, to automatically filter and score data extracted from OASST-formatted conversations, ensuring the curation of high-quality dialogue samples. The resulting datasets are structured under a flexible schema, enabling seamless integration into diverse training workflows. Together, these innovations offer a robust solution for generating and managing synthetic conversational data at scale, significantly reducing the overhead of data preparation in LLM training pipelines. Our code
and documentation are available at \url{https://github.com/ServiceNow/SyGra}
\end{abstract}


\section{Introduction}
\label{sec:intro}

The rapid progress of large language models (LLMs) and multimodal AI systems has heightened the demand for large-scale, high-quality training and evaluation datasets~\cite{brown2020language, touvron2023llama, openai2023gpt4}. Yet, the cost, bias, and limited availability of annotated real-world data present major barriers~\cite{yu2024makes}. This is especially true in areas like instruction tuning, tool-use supervision, multi-agent interactions, and safety evaluation, where fine-grained control over structure, diversity, and task complexity is essential~\cite{ouyang2022training, schick2023toolformer}.

Synthetic data, generated via LLMs and automated pipelines, offers greater flexibility and control than traditional datasets. Achieving this at scale, however, poses significant challenges: designing complex, branching workflows that mirror task hierarchies; orchestrating diverse model backends, APIs, and tool calls; enforcing validation and schema compliance across large, heterogeneous outputs; and enabling resumability, sharding, and streaming for scalable, fault-tolerant execution. Reusable, modular flows are also vital for maintainable pipelines.

For teams building domain-specific assistants—such as AI copilots, ticket triaging agents, or safety evaluators—these challenges lead to higher manual effort and slower iteration. A framework is needed that automates high-quality data generation, supports structured outputs and multimodal inputs, and streamlines augmentation—ultimately accelerating the development of custom LLMs for enterprise and research applications.

To address this, we introduce \textbf{\sygra} (\emph{Graph-oriented Synthetic-data Pipeline}), a general-purpose framework for scalable synthetic data generation. \sygra combines low-code, YAML-based configuration with modular, graph-driven orchestration to support complex workflows with branching, looping, and conditionals. It enables the reuse of graphs as subgraphs, ensures reliable execution through integrated validation and checkpointing, and natively supports multimodal inputs and agent based data generation. Additionally, \sygra offers unified dataset I/O across HuggingFace and local formats, supports quality tagging, and produces outputs compatible with OASST-style formatting for seamless downstream use.

\begin{table*}[!htbp]
\centering
\caption{Comparison of \sygra with popular frameworks across key capabilities.}
\label{tab:tool_comparison_grouped}
\resizebox{\textwidth}{!}{%
\begin{tabular}{llccccccc}
\toprule
\textbf{Category} & \textbf{Feature} & \textbf{\sygra} & \textbf{Distilabel} & \textbf{SDG} & \textbf{Curator} & \textbf{Synthetic Data Kit} \\
\midrule
\multirow{3}{*}{\textbf{Execution \& Authoring}} 
& Async Execution          & \checkmark & \checkmark & \checkmark & \checkmark & \checkmark \\
 
& Low-Code Authoring       & \checkmark & \xmark & \checkmark & \xmark & \checkmark \\
 
& UI-Based Flow Config     & \textcolor{orange}{$\triangle$} & \xmark & \checkmark & \checkmark & \xmark \\
 
\midrule
 
\multirow{2}{*}{\textbf{Workflow Orchestration}} 
& Configuration-driven Complex Flow & \checkmark & \checkmark & \textasteriskcentered & \checkmark & \textasteriskcentered \\
 
& Reusable Subgraphs & \checkmark & \xmark & \xmark & \xmark & \xmark \\
\midrule
 
\multirow{3}{*}{\textbf{Evaluation \& Integration}} 
& Quality Tagging         & \checkmark & \checkmark & \textasteriskcentered & \textasteriskcentered & \checkmark \\
 
& HuggingFace Integration & \checkmark & \checkmark & \checkmark & \checkmark & \checkmark \\
 
& Agent/Tool Support      & \checkmark & \checkmark & \xmark & \xmark & \xmark  \\
\midrule
\multirow{2}{*}{\textbf{Multimodality}} 
& Multimodal Input        & \checkmark & \textasteriskcentered & \xmark & \textasteriskcentered & \textasteriskcentered \\
& Multimodal Output       & \checkmark & \xmark & \xmark & \xmark & \xmark \\
\bottomrule
\end{tabular}
}
\begin{tablenotes}
\small
\centering
\item \checkmark: Supported \quad \xmark: Not Supported \quad \textcolor{orange}{$\triangle$}: Work in Progress \quad \textasteriskcentered: Partial Support
\end{tablenotes}
\end{table*}

\section{Related Work}
\label{sec:related_work}

Recent years have seen rapid progress in the development of synthetic data generation frameworks and instruction-tuning toolkits, with each system making distinct trade-offs across orchestration, extensibility, code abstraction, and multimodal support. Table \ref{tab:tool_comparison_grouped} summarizes core capabilities across representative frameworks like Distilabel\cite{distilabel2024}, SDG\cite{argilla_sdg2024}, Curator\cite{curator2024}, and Synthetic Data Kit\cite{llama_sdk2024}.

\begin{itemize}
    \item Existing data generation frameworks address only subsets of the end-to-end data generation pipeline, leaving gaps in orchestration, extensibility, and multimodal support.
    \item Most tools support some combination of asynchronous execution, low-code authoring, configuration-driven flows, and HuggingFace integration, but often lack reusable subgraphs, seamless UI-based workflow design, comprehensive agent/tool support, and integrated quality tagging.
    \item UI-based flow configuration is present in some tools (e.g., Curator), but these typically lack robust agent capabilities, multimodal I/O, or subgraphs.
\end{itemize}

In summary, while existing frameworks each offer valuable features for synthetic data generation, they typically address isolated aspects of the broader workflow. \sygra stands out by providing a unified, extensible approach that brings together the critical capabilities needed for modern, complex, and multimodal data generation pipelines.

\FloatBarrier
\begin{figure*}[!t]
  \centering
  \includegraphics[width=\textwidth]{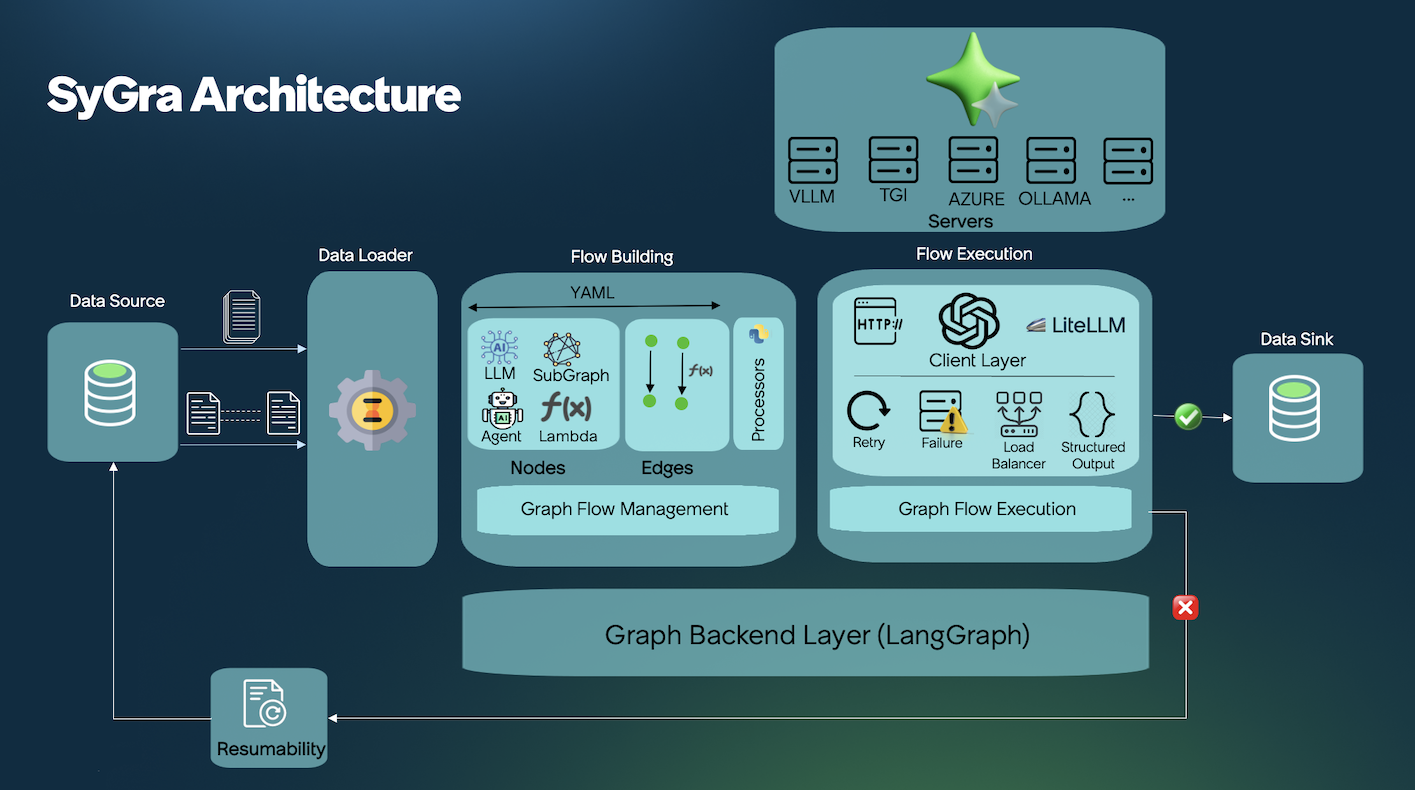}
  \caption{High-level \sygra architecture.}
  \label{fig:grasp_overview}
\end{figure*}

\section{\sygra Framework}
\label{sec:grasp_framework}

\sygra is a modular and extensible system designed for large-scale, programmable data generation. It supports configurable orchestration through a graph abstraction that enables reusable, auditable, and resumable workflows. The framework is designed for both research and production pipelines, with pluggable model backends and modular task authoring support.

\subsection{System Architecture}

\sygra is guided by three principles—\textbf{Scalability} (streaming data sources, resumable jobs, JSONL/Parquet/HF outputs), \textbf{Modularity} (YAML-defined DAG workflows with conditional logic), and \textbf{Reusability} (versioned, reusable graphs, nodes, and validators). Figure~\ref{fig:grasp_overview} shows its core components:

\begin{enumerate}
  \item \textbf{Data I/O}: Unified loader/sink for HuggingFace or local CSV, JSON(L), and Parquet in batch or streaming modes.

  \textit{Notably, \sygra provides support for ServiceNow as a data source and sink}, enabling seamless integration with ServiceNow instances~\cite{servicenow_api}.

  \item \textbf{Graph Construction}: YAML-defined DAG of nodes (LLM calls, transformations) with conditional edges and pre/post hooks, compiled via LangGraph~\cite{langgraph2024}
  \item \textbf{Execution Engine}: Asynchronous runtime coordinating local Python steps and remote inference (HTTP, OpenAI, Mistral) across VLLM~\cite{kqiao2023vllm}, TGI~\cite{tgi2023}, OLLAMA~\cite{ollama2024}, and Azure/OpenAI backends~\cite{openai2023gpt4}, with built-in retries and failure tracing.
  \item \textbf{Structured Output \& Resumability}: Generates OASST-compatible\cite{kopf2023openassistant} records and tracks progress metadata for fault-tolerant, restartable runs.
\end{enumerate}

\subsection{Pipeline Components}
\label{subsec:pipeline_components}

\sygra pipelines are defined declaratively in YAML, promoting low-code, reproducible workflow construction. Each pipeline consists of three configuration blocks:

\paragraph{Data Configuration (\texttt{data\_config}).}
Specifies input and output sources, format handling (CSV, JSONL, Parquet), streaming options, and inline preprocessing (e.g., renaming, filtering, and combining). Supports both data-backed and data-less generation scenarios.

\begin{figure}[!htbp]
    \centering
    
    \begin{minipage}{0.47\textwidth}
        \centering
        \fcolorbox{black}{white}{\includegraphics[width=\linewidth]{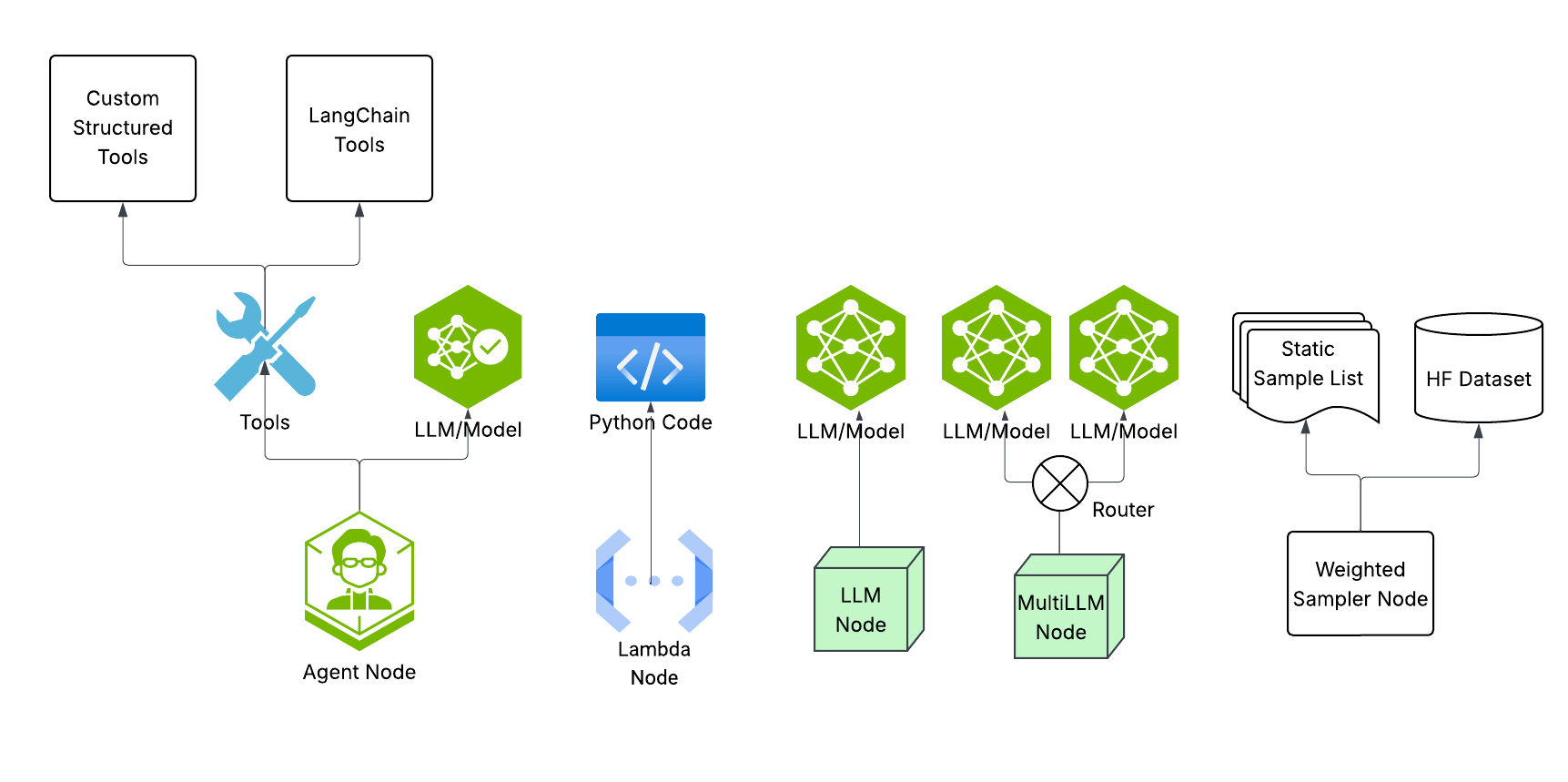}}
        \caption{Graph nodes}
        \label{fig:graph-nodes}
    \end{minipage}
    \hfill
    \begin{minipage}{0.47\textwidth}
        \centering
        \fcolorbox{black}{white}{\includegraphics[width=\linewidth]{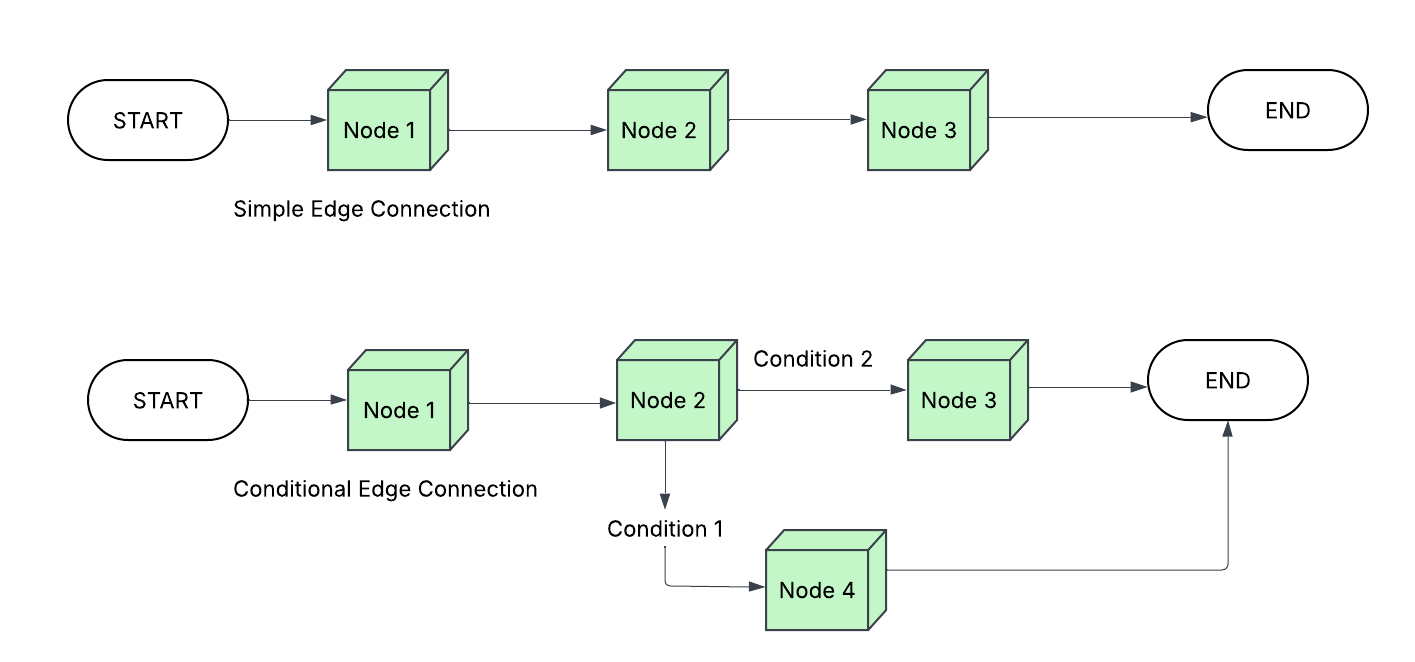}}
        \caption{Graph edges}
        \label{fig:graph-edges}
    \end{minipage}

\end{figure}

\paragraph{Graph Configuration (\texttt{graph\_config}).}
Defines a DAG of computational nodes (Figure \ref{fig:graph-nodes}), each node can be configured to call
LLMs, Python functions, agents, or subgraphs. Various node types are supported, such as: 
\begin{itemize}
    \item \textbf{\texttt{llm}}: When a model needs to be called, we can use a LLM node model properties and role-based prompts, along with pre/post processors in Python code. 
    \item \textbf{\texttt{multi\_llm}}: When we need to generate data at scale, we can use a multi-LLM node which allows configuration of load balanced model inferences between multiple endpoints. 
    \item \textbf{\texttt{lambda}}: To process the data during execution, we can utilize lambda nodes, which are mapped to Python functions.
    \item \textbf{\texttt{agent}}: To perform end-to-end agentic behaviour, we can use agent nodes along with tools, which can be custom or Langchain tools.
    \item \textbf{\texttt{subgraph}}: Complex flows can be splitted into smaller graphs i.e. subgraphs which can be reused inside the graph.
\end{itemize}

Once the nodes are defined, we connect them via egdes (Figure \ref{fig:graph-edges}). Two types are supported -- simple edge and conditional edge. Conditional edges are useful to build if-else flow and loops in the graph based on a condition written as a Python code. 

\paragraph{Output Configuration (\texttt{output\_config}).}
Controls how graph states are serialized into structured output. Users can declaratively map, transform, or customize output using Python hooks to match target schemas like OASST.

\paragraph{Schema Validation.}
Ensures output integrity via type and rule-based validation. Schemas can be defined in YAML or Python (e.g., Pydantic), with invalid records automatically skipped and logged.

Finally, the graph is validated and compiled into a LangGraph-compatible representation. Refer to Appendix~\ref{app:pipeline_config} for detailed configuration options and schema definitions.

\subsection{Key Features}

\sygra brings together robust design abstractions and practical scalability for real-world use cases. Specifically, our contributions include:
\begin{enumerate}
  \item \textbf{Low-Code, Modular Graph Configuration:} \sygra combines a YAML-based interface with LangGraph-style agents and a custom DAG engine, enabling concise, extensible definitions of complex workflows with branching, looping, and conditionals.~\ref{app:graph_config_example}

  \begin{lstlisting}[style=yaml, frame=single]
    data_config:
      source:
        type: "hf"
        repo_id: "google-research-datasets/mbpp"
        config_name: "sanitized"
        split: ["train"]
    graph_config:
      nodes:
        generate_answer:  
          node_type: llm        
          prompt:
            - system: |
                You are an assistant tasked with solving python coding problems. 
            - user: |
                {prompt}      
          model:      
            name: gpt-4o
            parameters:         
              temperature: 0.1
      # more nodes defined here like critique answer
      edges:
        - from: START
          to: generate_answer
        - from: generate_answer
          to: critique_answer
        - from: critique_answer
          to: END
    output_config:
      output_map:
        id:
          from: "id"
        conversation:
          from: "messages"
  \end{lstlisting}
  \item \textbf{Reusable Recipes (Subgraphs):} This feature enables us to use common graph components which can be reused across tasks, promoting modularity. For instance, the Evolve INSTRUCT recipe (Figure ~\ref{fig:instruction_evolver_flow}) encapsulates a modular subgraph that receives seed instructions and applies either depth-based or breadth-based evolution strategies via a routing node (Strategy)~\cite{wizardlm}. This subgraph can be invoked repeatedly across different flows, enhancing composability and reducing redundancy.

  \begin{figure}[!htbp]
    \centering
    \includegraphics[width=0.9\linewidth]{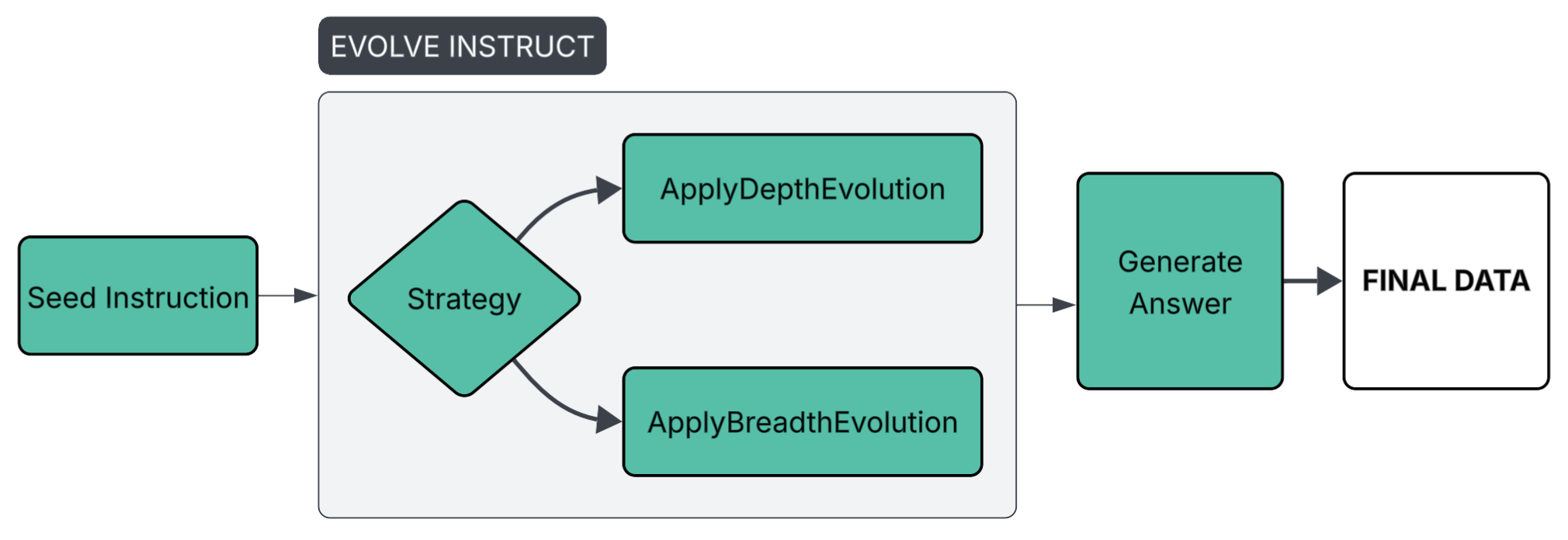}
    \caption{Instruction evolution subgraph and judgment loop used within \sygra pipelines.}
    \label{fig:instruction_evolver_flow}
  \end{figure} 
  
  \item \textbf{Multimodal Support:} \sygra extends beyond text-only workflows by natively handling \textbf{audio} and \textbf{image} inputs alongside text. Through unified I/O adapters, it transparently loads local or remote media in various formats, encodes them as base64 data URLs for LLM API compatibility, and supports multiple media fields per record. This enables workflows for tasks such as speech recognition, audio classification, document analysis, and visual QA. Round-tripping ensures outputs can be saved back into HuggingFace datasets in their original formats for reproducibility and downstream use. Additionally, \sygra supports multimodal outputs—including generated images and audio—when using GPT-based endpoints (OpenAI or Azure OpenAI), enabling end-to-end multimodal generation for OpenAI endpoints.

  \begin{lstlisting}[style=yaml, frame=single]
    identify_animal:
      output_keys: animal
      node_type: llm
      prompt:
        - user:
            - type: text
              text: |
                Identify the animal in the provided audio.
            - type: audio_url
              audio_url: "{audio}"

      model:
        name: qwen_2_audio_7b
        parameters:
          max_tokens: 1000
          temperature: 0.3
  \end{lstlisting}
  
  \item \textbf{Agentic Execution:} \sygra enables the creation of autonomous, tool-using agents built on the ReAct reasoning-and-acting paradigm via LangGraph. Agent nodes extend LLM nodes with capabilities for dynamic tool invocation, multi-turn reasoning, and conditional decision-making. Developers can specify a library of callable tools, inject context-specific system messages at arbitrary conversation turns, and configure pre/post-processing hooks for fine-grained control over input and output. This allows pipelines to handle exploratory tasks, iterative search, and interactive decision flows in a modular, low-code manner.

\begin{lstlisting}[style=yaml, frame=single]
  research_agent:
    node_type: agent
    prompt:
      - system: |
          You are a research assistant that helps users find information.
          Always think step by step and explain your reasoning.
      - user: |
          Please help me research {topic}.
    tools:
      - tasks.sim.tools.search_tool.search
      - tasks.sim.tools.calculator_tool.calculate
    inject_system_messages:
      2: "Remember to cite your sources."
    output_keys:
      - agent_response
    model:
      name: vllm_model
      parameters:
        temperature: 0.2
        max_tokens: 1024
\end{lstlisting}

  \item \textbf{Structured Output Generation:} \sygra provides a flexible framework for generating and validating \emph{structured outputs} from LLMs, reducing post-processing effort and ensuring reliable formats. It supports both \textbf{class-based schemas} (via Pydantic) and \textbf{YAML-defined schemas}, with automatic type handling and optional custom validation rules. Structured output generation works natively with \texttt{OpenAI} and \texttt{vLLM} models, and falls back to JSON schema validation for other backends. This allows developers to define precise field types, attach descriptions, and enforce constraints directly at generation time.

\begin{lstlisting}[style=yaml, frame=single]
  nodes:
    answer_node:
      node_type: llm
      model:
        name: gpt-4o
        parameters:
          temperature: 0.1
      structured_output:
        enabled: true
        schema:
          fields:
            answer:
              type: str
              description: "Main answer text"
            confidence:
              type: float
              description: "Confidence score between 0 and 1"
\end{lstlisting}
  
  \item \textbf{Resumability:} \sygra supports fault-tolerant, restartable execution of long-running jobs. In the event of a failure, execution can gracefully shut down and later resume from the last recorded checkpoint without reprocessing completed steps. This is particularly valuable for large-scale or streaming workloads where partial progress should be preserved. Checkpoints store both intermediate outputs and node-level metadata, enabling accurate restoration of execution state.

    \begin{lstlisting}[language=bash, frame=single]
    python main.py --task <your_task> --resume True
    \end{lstlisting}
  \item \textbf{Metadata Tracking:} \sygra includes an automatic metadata tracking system that captures comprehensive execution metrics without requiring any code changes. The system provides real-time cost tracking for multiple LLM providers (OpenAI, Azure OpenAI, Anthropic Claude on AWS Bedrock, vLLM), detailed token usage statistics, and multi-level performance monitoring at aggregate, model, and node granularities. Metrics include latency percentiles (p50, p95, p99), throughput measurements, retry and failure rates, and response code distributions. Output and metadata files share synchronized timestamps for easy correlation, and the system captures execution context including git commit information and dataset versioning for full reproducibility.

  \begin{lstlisting}[language=Python, frame=single, showstringspaces=false]
    from sygra.metadata.metadata_collector import get_metadata_collector

    collector = get_metadata_collector()
    metadata = collector.get_metadata_summary()

    # Access aggregate metrics
    stats = metadata['aggregate_statistics']
    print(f"Total cost: ${stats['cost']['total_cost_usd']:.4f}")
    print(f"Total requests: {stats['requests']['total_requests']}")
    print(f"Models used: {list(metadata['models'].keys())}")
    print(f"Nodes executed: {list(metadata['nodes'].keys())}")
  \end{lstlisting}
  
  \item \textbf{Filterable OASST-Compatible Formatting:} Outputs can be structured in an OASST-compatible~\cite{kopf2023openassistant} format for easy post-hoc filtering, inspection, and training integration.
  
  \begin{figure}[!htbp]
    \centering
    \includegraphics[width=0.9\linewidth]{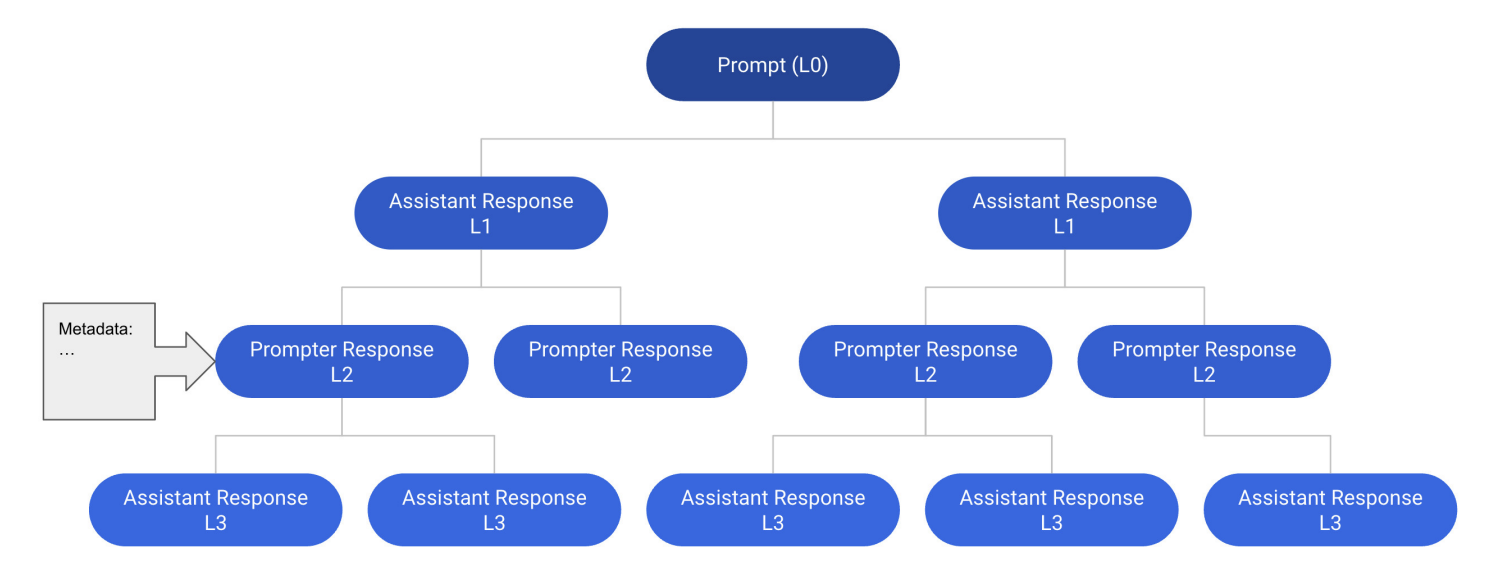}
    \caption{An example Conversation Tree of depth 4 containing 12 messages~\cite{kopf2023openassistant}}
    \label{fig:oasst}
  \end{figure} 

\end{enumerate}

\section{Dual-Stage Quality Tagging}
\label{sec:quality}

Quality control is central to synthetic data generation. \sygra implements a two-stage mechanism balancing efficiency and accuracy: fast heuristic filtering eliminates obvious low-quality samples, followed by targeted LLM-based evaluation for samples passing initial checks. This section details both stages, metadata schema, and integration with training pipelines.

\subsection{Stage 1: Heuristic-Based Filtering}

The first stage applies eight checks. Implementation uses thread pool with $N_{cpu}$ workers.

\textbf{1. Conversation Pretokenization:} Applies model-specific chat template (fetched from HuggingFace or custom) to validate format compatibility. Computes token count using tokenizer. 

\textbf{2. Language Detection:} Uses fastText~\cite{joulin2017bag} for detection (99.5\% accuracy on XLM-R benchmark). Concatenates all turns, detects language, computes confidence. Rejects if detected language not in target set (default: \{English\}) or confidence $<$ threshold (default: 0.90). Useful for: filtering code-switched data, enforcing monolingual datasets, handling web-scraped data with mixed languages.

\textbf{3. Conversation Length Check:} Counts turns and validates range. Rejects if: (a) turns $<$ \texttt{min\_turns} (default: 2, requires at least one exchange), or (b) turns $>$ \texttt{max\_turns} (default: 20, filters extremely long dialogues that are often errors or edge cases). Prevents: empty conversations, infinitely long generations (from LLM loops), single-turn data in multi-turn pipelines.

\textbf{4. Metadata Tagging:} Extracts structural features for downstream filtering and analysis: turn count, average turn length (chars), role distribution (\% assistant vs. user), special token usage (code blocks, math symbols, citations). Stored in metadata but does not reject samples. Used later for: stratified sampling by length, balancing role distributions, identifying domain-specific data.

\textbf{5. Lexical Diversity:} Computes Type-Token Ratio (TTR) and Measure of Textual Lexical Diversity (MTLD)~\cite{mccarthy2010mtld}:
\begin{align}
\text{TTR} &= \frac{|\text{unique tokens}|}{|\text{total tokens}|} \\
\text{MTLD} &= \frac{1}{k} \sum_{i=1}^{k} \ell_i \quad \text{where } \ell_i \text{ is length to factor $<$0.72}
\end{align}
Rejects if TTR $<$ 0.30 (highly repetitive text, e.g., "I like cats. I like dogs. I like birds."). MTLD captures diversity over longer spans, complementing TTR's sensitivity to sample length. Filters: LLM generation loops (same phrases repeated), template-based spam, low-effort responses.

\textbf{6. Perplexity Scoring:} Uses domain-specific language model (default: GPT-2 small for speed) to compute perplexity:
\begin{equation}
\text{PPL}(x) = \exp\left(-\frac{1}{N}\sum_{i=1}^{N} \log P(x_i | x_{<i})\right)
\end{equation}
Rejects if PPL $>$ threshold (default: 1000, indicating low fluency). Catches: grammatically malformed text, gibberish, foreign languages not caught by detector, heavily code-mixed data. Trade-off: requires GPU for fast scoring (CPU: 50 samples/sec, GPU: 500 samples/sec). Optional—users can disable if throughput is critical.

\textbf{7. Reward Modeling (Optional):} Applies pre-trained reward model to predict human preference score. Rejects if score $<$ threshold (default: 0.3). Provides semantic quality signal beyond surface features. Trade-off: expensive (requires GPU, 100-200ms per sample). Recommended for high-value datasets where cost is acceptable.

\textbf{8. Data Characteristics:} Computes domain-specific features used for routing and analysis: presence of code blocks, mathematical symbols, citations, tables, lists. Determines conversation category (generic, math, code, QA). Stored in metadata for Stage 2 routing. No rejection—purely informational.


\subsection{Stage 2: LLM-Based Categorical Evaluation}

Samples after Stage 1 undergo category-specific LLM evaluation. 


\textbf{Category Classification:} Based on Data Characteristics output from Stage 1, samples are classified into seven categories:
\begin{itemize}[noitemsep]
    \item \textbf{generic}: General conversation, no specific domain
    \item \textbf{math\_solving}: Mathematical reasoning, problem-solving
    \item \textbf{reasoning}: Logical reasoning, deduction, inference
    \item \textbf{code\_writing}: Programming, code generation/explanation
    \item \textbf{complex\_instruction\_following}: Multi-step instructions, constraints
    \item \textbf{open\_qa}: Open-domain question answering, factual queries
    \item \textbf{closed\_qa}: Closed-domain QA, specific context provided
\end{itemize}

Classification uses simple rules (presence of code blocks $\rightarrow$ code\_writing, math symbols $\rightarrow$ math\_solving, etc.) with GPT-4o fallback for ambiguous cases.

\textbf{Evaluation Dimensions:} Each category evaluator assesses five dimensions on 1-5 scale:
\begin{itemize}[noitemsep]
    \item \textbf{Instruction Following (IF):} Adherence to constraints, completeness of response relative to request
    \item \textbf{Contextual Alignment (CA):} Relevance to conversation history, appropriate continuations
    \item \textbf{Accuracy (AC):} Factual correctness, logical validity, absence of hallucinations
    \item \textbf{Completeness (CO):} Thoroughness, addressing all aspects of query
    \item \textbf{Linguistic Clarity (LC):} Grammar, coherence, readability, fluency
\end{itemize}

\textbf{Category-Specific Prompts:} Each category uses specialized evaluation prompt optimized for relevant quality aspects. For example:
\begin{itemize}[noitemsep]
    \item \textbf{math\_solving}: Emphasizes step-by-step correctness, formula accuracy, numerical precision
    \item \textbf{code\_writing}: Focuses on syntax validity, logic correctness, edge case handling, efficiency
    \item \textbf{reasoning}: Evaluates logical coherence, premise-conclusion validity, absence of fallacies
\end{itemize}

Prompts include few-shot examples for calibration (2-3 examples per category).

\textbf{Output Format:} Evaluators return structured JSON:
\begin{lstlisting}[language=JSON, frame=single]
{
  "instruction_following": 4,
  "explanation_IF": "Response addresses all...",
  "contextual_alignment": 5,
  "explanation_CA": "Perfectly aligned with...",
  "accuracy": 4,
  "explanation_AC": "Factually correct but...",
  "completeness": 4,
  "explanation_CO": "Covers main points...",
  "linguistic_clarity": 5,
  "explanation_LC": "Clear, well-structured..."
}
\end{lstlisting}

JSON enforcement via model's structured output mode (GPT-4o, Claude 3.5) or constrained decoding (vLLM with guidance). Explanations provide interpretability and debugging support.


\subsection{Metadata Schema and Integration}

Quality metadata follows hierarchical structure compatible with OASST format:
\begin{lstlisting}[language=JSON]
{
  "conversation": [...],  // original dialogue
  "metadata": {
    "quality_characteristics": {
      "heuristic_based": {
        "lexical_richness": {
          "ttr_score": 0.67,
          "mtld_score": 89.3
        },
        "perplexity": {"score": 234.5},
        "language": {
          "detected": "en",
          "confidence": 0.98
        },
        "conversation_stats": {
          "turn_count": 4,
          "avg_turn_length": 127
        }
      },
      "LLM_based": {
        "category": "math_solving",
        "instruction_following": 4,
        "contextual_alignment": 5,
        "accuracy": 4,
        "completeness": 4,
        "linguistic_clarity": 5,
        "explanations": {...}
      }
    }
  }
}
\end{lstlisting}

This format integrates seamlessly with training pipelines:
\begin{itemize}[noitemsep]
    \item \textbf{Filtering:} Reject samples where any dimension $<$ threshold (e.g., accuracy $<$ 3 for factual datasets)
    \item \textbf{Stratified Sampling:} Balance quality distributions (e.g., equal numbers of scores 3, 4, 5)
    \item \textbf{Reward Modeling:} Use dimension scores as auxiliary supervision signals
    \item \textbf{Curriculum Learning:} Order training samples by difficulty (ascending average score)
    \item \textbf{Weighted Sampling:} Sample probability proportional to quality score during training
\end{itemize}

OASST compatibility enables direct use with HuggingFace Transformers' SFT and DPO trainers without format conversion.

\section{Results and Impact}

\subsection{Experimental Setup}

The evaluation was run on an \textbf{8-core CPU machine with 16 GB RAM} using the \textbf{\sygra framework}. Model endpoints were deployed separately on \textbf{vLLM} with the \textbf{Qwen 3 32B Instruct} model, configured to use \textbf{two GPUs with tensor parallelism}, moderate CPU and memory resources, and optimizations such as \textbf{chunked prefill} with capped GPU memory usage.

The workload consisted of \textbf{10,000 input records}, repeated across three trials at each concurrency level. To keep the focus on system behavior, the workflow was deliberately simple: a \textbf{weighted sampler node} selected tone and persona values, and an \textbf{LLM node} rephrased the input text using fixed generation parameters (deterministic outputs, max length 500 tokens). This design provided a controlled inference workload for measuring concurrency scaling without pipeline complexity.

\subsection{Results}

Performance was measured in terms of \textbf{total wall-clock time} required to process the workload under varying concurrency levels (denoted here as \emph{batch size}).

\begin{figure}[h]
    \centering
    \includegraphics[width=1\linewidth]{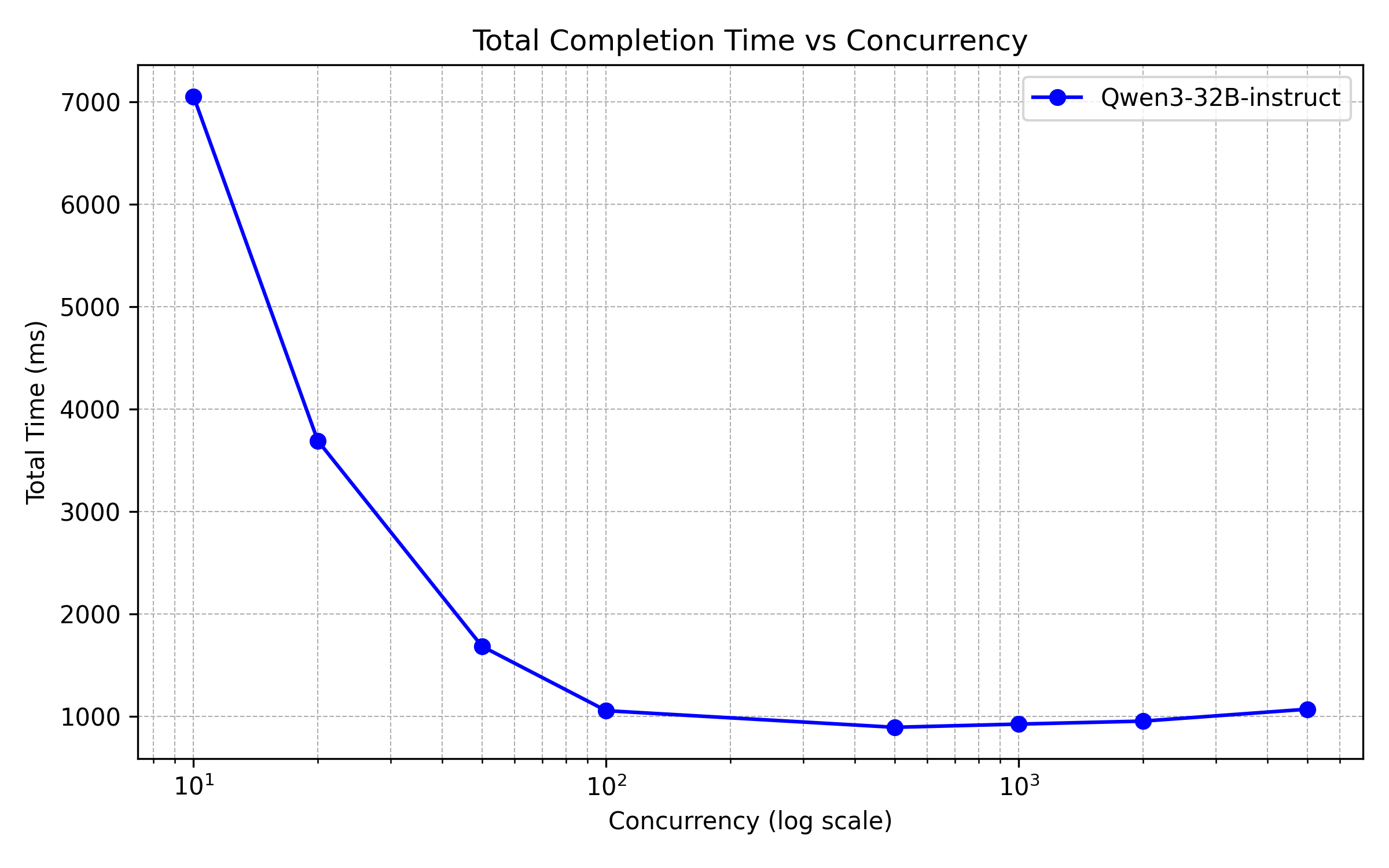}
    \caption{Total completion time (seconds) for varying concurrency levels with one endpoint.}
    \label{fig:grasp_scaling}
\end{figure}

Three key patterns emerge:
\begin{itemize}
    \item \textbf{Reduced completion time with higher concurrency.} As concurrency increases from 10 to $\sim$500, total completion time decreases sharply, illustrating the efficiency of \sygra’s asynchronous execution.
    \item \textbf{Sustained efficiency at scale.} Beyond $\sim$500 concurrent requests, completion time stabilizes around 900--1000 seconds, showing resilience under heavy load.
    \item \textbf{Server saturation at very high concurrency.} At batch size 5000, total time increases slightly. This reflects a limitation of the underlying model server rather than \sygra itself. Users can mitigate this by \textbf{adding multiple endpoints} or \textbf{increasing the computational resources allocated to the hosted model}.
\end{itemize}

\textbf{Note on Task Complexity.} The experiment used a deliberately simple two-node workflow (sampling + rephrasing) to isolate concurrency effects. In this setting, a single vLLM endpoint was sufficient to handle high concurrency. However, in \textbf{more complex tasks with multiple nodes and richer pipelines}, \sygra exhibits stronger scaling when additional endpoints or resources are available, as state-of-the-art inference servers such as vLLM are capable of sustaining concurrent requests efficiently across distributed deployments.

\subsection{Impact}

These results underscore \sygra’s ability to deliver \textbf{stable, predictable performance under extreme concurrency}. Whereas traditional multi-threaded or multi-process inference servers often degrade sharply once concurrency exceeds a few hundred requests, \sygra sustains bounded completion times even at \textbf{$>$5000 concurrent requests}.

The implications are significant:
\begin{itemize}
    \item \textbf{Scalable deployments.} \sygra enables operators to support thousands of simultaneous queries while maintaining consistent workload completion times.
    \item \textbf{Configurable performance.} If saturation is observed at very high concurrency, performance can be extended by scaling to multiple endpoints or allocating additional resources to the hosted model.
    \item \textbf{Future readiness.} As LLM-based systems (e.g., conversational agents, retrieval-augmented generation) face growing concurrency demands, \sygra provides a reliable inference layer that avoids bottlenecks at production scale.
\end{itemize}

In summary, \sygra demonstrates \textbf{robust concurrency scaling, endpoint-agnostic performance, and stable execution times}, making it a strong candidate for next-generation large-scale data syntheses.

\section{Availability}
\sygra is released as an open-source Python package and framework on PyPI\footnote{\texttt{pip} package: \url{https://pypi.org/project/sygra/}} (\texttt{pip install sygra}) and GitHub\footnote{https://github.com/ServiceNow/SyGra}. 

\section{Conclusion}
\label{sec:conclusion}

We presented \textbf{\sygra}, a modular framework for synthetic data generation using graph-based, prompt-centric workflows. \sygra offers scalable, reproducible pipelines for language model training, featuring a low-code YAML interface, reusable subgraphs, agent nodes, and HuggingFace-native I/O. Its design supports diverse workflows, uniquely enabling multimodal inputs, subgraph reuse, conditional routing, and schema validation.

Current limitations include multimodal outputs being restricted to GPT-based endpoints (other backends remain text-only), independent node operation without cross-sample reasoning, and basic agent support.

\sygra accelerates dataset creation and promotes transparency and reuse in LLM development. Ongoing efforts must address risks like “model collapse” through mixed datasets and continuous quality control, ensuring \sygra’s utility across generative AI applications.

\section*{Acknowledgements}
We gratefully acknowledge the following people for their contributions: Nirali Popat, Sidharthenee Nayak, Nandhakumar Kandasamy, Sravan Ramachandran, Segan Subramanian, Masoud Hashemi and Rishabh Maheshwary.


{\small
\bibliographystyle{abbrvnat}
\bibliography{egbib}
}

\clearpage
\appendix



\section{Pipeline Components: Features, Definitions and Examples}
\label{app:pipeline_config}

\subsection{Data Configuration}
\label{app:data_config_example}
\subsubsection{Input Sources} 
\noindent

\noindent
This configuration illustrated below \ref{lst:data-config} represents:
\begin{itemize}
  \item Input from HuggingFace and local disk (alternative)
  \item Use of \texttt{RenameFieldsTransform} for renaming schema fields
  \item Optional sink setup with HuggingFace or local file export
\end{itemize}

\begin{lstlisting}[style=yaml, frame=single, title={An example configuration using a HuggingFace dataset as source and applying field renaming transformation is shown below.}, label={lst:data-config}]
data_config:
  source:
    # Example 1: HuggingFace dataset source
    type: "hf"                               # HuggingFace dataset
    repo_id: "google-research-datasets/mbpp" # HuggingFace repository ID
    config_name: "sanitized"                 # Dataset configuration name
    split: ["train", "validation", "prompt"] # Dataset splits to use

    # OR

    # Example 2: Local file source
    type: "disk"                             # Local file source
    file_path: "/path/to/data.json"          # Path to input file
    file_format: "json"                      # Format (json, jsonl, csv, parquet)
    encoding: "utf-8"                        # File encoding

    # Optional transformations to apply to the input data
    transformations:
      - transform: sygra.processors.data_transform.RenameFieldsTransform  # Path to transformation class
        params:                                                     # Parameters for the transformation
          mapping:
            task_id: id                     # Rename 'task_id' field to 'id'
          overwrite: false                  # Don't overwrite existing fields
          
  # Optional sink configuration for where to store output data
  sink:
    # Example 1: HuggingFace dataset sink
    type: "hf"                               # HuggingFace dataset
    repo_id: "output-dataset/synthetic-mbpp" # Where to upload the data
    split: "train"                           # Split to write to
    private: true                            # Create a private dataset
    
    # OR
    
    # Example 2: Local file sink
    type: "json"                             # File format (json, jsonl, csv, parquet)
    file_path: "/path/to/output/file.json"   # Path to save the file
    encoding: "utf-8"                        # File encoding
\end{lstlisting}

\subsubsection{Transformations} 

\paragraph{RenameFieldsTransform.} The \texttt{RenameFieldsTransform} is a lightweight transformation utility used in the \sygra pipeline to rename one or more fields in each record of the dataset. This is particularly useful for ensuring consistency in variable naming, aligning raw data to prompt-ready formats, or preparing input fields for downstream processing.

The YAML configuration for this transformation accepts a \texttt{mapping} parameter, which specifies how input field names should be renamed. An optional \texttt{overwrite} flag determines whether to overwrite any existing field in case of name collision.

Example below shows a sample usage where the fields \texttt{page}, \texttt{llm\_extract}, and \texttt{type} are renamed to \texttt{id}, \texttt{text}, and \texttt{text\_format}, respectively.

\begin{lstlisting}[style=yaml, frame=single, title={Example usage of RenameFieldsTransform in YAML configura-
tion. This renames selected fields to align with graph input expectations.}, label={lst:rename_fields_transform}]
  - transform: sygra.processors.data_transform.RenameFieldsTransform
    params:
      mapping:
        page: id
        llm_extract: text
        type: text_format
\end{lstlisting}






\paragraph{CombineRecords.} This transformation combines multiple records to form richer contextual input. It can skip from the beginning or end of the dataset, define how many records to combine, and how to shift the combination window. As shown below, the configuration merges two records, joining multiple fields with newline delimiters or preserving the first record's values.

\begin{lstlisting}[style=yaml, frame=single, label={lst:combine_records}]
  - transform: sygra.processors.data_transform.CombineRecords
    params:
      skip:
        from_beginning: 10
        from_end: 10
      combine: 2
      shift: 1
      join_column:
        page: "$1-$2"
        pdf_reader: "$1\n\n$2"
        llm_extract: "$1\n\n$2"
        type: "$1"
        model: "$1"
        metadata: "$1"
\end{lstlisting}

\paragraph{SkipRecords.} It presents a simpler configuration to exclude records from the dataset, either from the start or end. This is especially useful for filtering noisy, incomplete, or structurally incompatible entries prior to processing.

\begin{lstlisting}[style=yaml, frame=single, label={lst:skip_records}]
  - transform: sygra.processors.data_transform.SkipRecords
    params:
      skip_type: "count"
      count:
        from_start: 10
        from_end: 10
\end{lstlisting}

\subsubsection{Data Less Mode} 

In data-less mode, \sygra operates without any input source. Instead, it directly executes the graph and writes outputs based solely on intermediate or generated values. This is especially useful for bootstrapping datasets, performing zero-shot synthesis, or generating instructional data.

The below YAML shows a minimal configuration that defines only an output sink.

\begin{lstlisting}[style=yaml, frame=single, label={lst:data_less}]
data_config:
  # No source configuration
  
  # Only sink configuration
  sink:
    type: "json"
    file_path: "output/synthetic_data.jsonl"
\end{lstlisting}

\subsection{\texttt{graph\_config}: Nodes and Execution Flow}

\textbf{Graph-Level Properties:}
\begin{itemize}
  \item \texttt{chat\_conversation: singleturn} or \texttt{multiturn}
  \item \texttt{chat\_history\_window\_size}: integer
\end{itemize}

\textbf{Node Types:}
\begin{itemize}
  \item \texttt{llm} — standard prompt inference
  \item \texttt{multi\_llm} — ensemble-style multi-model generation
  \item \texttt{weighted\_sampler} — controlled randomness
  \item \texttt{lambda} — run Python logic
  \item \texttt{agent} — multi-turn agent execution with memory and tools
  \item \texttt{subgraph} — reusable logical block
\end{itemize}

Each node can define:
\begin{itemize}
  \item Prompt templates with variable substitution
  \item Model name and parameters
  \item Input/output keys, chat history, role labeling
  \item Pre-process and post-process functions
\end{itemize}

\textbf{Edge Types:}
\begin{itemize}
  \item \textbf{Simple Edges:} Direct transitions between nodes.
  \item \textbf{Conditional Edges:} Conditional routing via Python classes and \texttt{path\_map}.
\end{itemize}

Special nodes: \texttt{START} and \texttt{END} are implicit entry and exit points.

\subsection{\texttt{output\_config}: Record Generation}

\textbf{Declarative Output Mapping:}
Each field in \texttt{output\_map} can use:
\begin{itemize}
  \item \texttt{from}: Reference a graph state variable
  \item \texttt{value}: Assign a static constant
  \item \texttt{transform}: Apply method in generator class
\end{itemize}

Supports context-aware templating with \texttt{\$} paths to inject YAML metadata (e.g., \texttt{\$data\_config.source.repo\_id}).

\vspace{0.5em}
\textbf{Custom Output Generators:}
Advanced logic can override the \texttt{generate()} method to control formatting or field post-processing.

\subsection{\texttt{schema\_config}: Output Validation}

\sygra supports both declarative and programmatic schema enforcement:

\textbf{Option 1: YAML-based Schema}
\begin{itemize}
  \item Define \texttt{fields} with name, type, and optional rules (e.g., \texttt{is\_greater\_than}, \texttt{regex}).
\end{itemize}

\textbf{Option 2: Python Schema Class}
\begin{itemize}
  \item Define a class extending \texttt{BaseModel}, use Pydantic \texttt{@validator} or \texttt{@root\_validator}.
\end{itemize}

Validation is applied post-execution; failing records are logged and skipped.
\\[0.4cm]

\begin{lstlisting}[style=yaml, frame=single]
  # Example A: use a custom Pydantic schema class
  schema_config:
    schema: validators.custom_schemas.CustomUserSchema
\end{lstlisting}

\begin{lstlisting}[style=yaml, frame=single]
  # Example B: inline field schema with rules
  schema_config:
    fields:
      - name: id
        type: int
        is_greater_than: 99999   # ensure >= 6 digits
      - name: conversation
        type: list[dict[str, any]]
      - name: taxonomy
        type: list[dict[str, any]]
      - name: annotation_type
        type: list[str]
      - name: language
        type: list[str]
      - name: tags
        type: list[str]
\end{lstlisting}

\subsection{Post-Generation Extensions}

\textbf{OASST Mapper:}
Enables conversion of records into SFT/DPO format based on the OpenAssistant schema. Activate with: \texttt{--oasst True}


\textbf{Quality Tagging:}
Automatically tags records using LLMs or heuristics. Enable with: \texttt{--quality True}


\section{Example \sygra YAML Configurations}
\label{app:graph_config_example}

This appendix provides example YAML configurations illustrating how \sygra pipelines are defined and composed using the \texttt{data\_config}, \texttt{graph\_config}, \texttt{output\_config}, and \texttt{schema\_config} sections. 
These examples demonstrate \sygra's flexibility for data-driven and zero-shot pipelines, LLM orchestration, and safe output generation.

\subsection{Minimal Data-Less Generation Configuration}

\begin{lstlisting}[style=yaml, frame=single]
    data_config:
      sink:
        type: "json"
        file_path: "output/synthetic_data.jsonl"
    
    graph_config:
      nodes:
          generate:
            node_type: llm
            output_keys: response
            prompt:
              - system: "You are a helpful assistant."
              - user: "Write a fun fact about space."
            model:
              name: gpt-3.5-turbo
              parameters:
                temperature: 0.8
      edges:
        - from: START
          to: generate
        - from: generate
          to: END
    
    output_config:
      output_map:
      fact:
        from: response
\end{lstlisting}

\subsection{Full Pipeline with Conditional Edge and Schema Validation}

\begin{lstlisting}[style=yaml, frame=single]
    data_config:
      source:
        type: "disk"
        file_path: "data/code_tasks.jsonl"
        file_format: "jsonl"
      sink:
        type: "jsonl"
        file_path: "output/validated_output.jsonl"
    
    graph_config:
      nodes:
        generate:
          node_type: llm
          output_keys: solution
          prompt:
            - system: "You are an AI that solves code problems."
            - user: "{task}"
          model:
            name: mistral
            parameters:
              temperature: 0.5
        validate:
          node_type: lambda
          lambda: validators.code.check_validity
          output_keys: 
            - is_valid
      edges:
        - from: START
          to: generate
        - from: generate
          to: validate
        - from: validate
          condition: validators.code.RouteBasedOnValidity
          path_map:
            END: END
            generate: generate
    
    output_config:
      output_map:
        id:
          from: task_id
        solution:
          from: solution
        validity:
          from: is_valid
    
    schema_config:
      fields:
        - name: id
          type: int
        - name: solution
          type: str
        - name: validity
          type: bool
\end{lstlisting}

\subsection{Pipeline to process images as input}

\begin{lstlisting}[style=yaml, frame=single]
    data_config:
      source:
        type: "hf"
        repo_id: "datasets-examples/doc-image-1"
        split: "train"
        streaming: true
    
      sink:
        type: "hf"
        repo_id: <repo_name>
        config_name: MM-doc-image-1
        split: train
        push_to_hub: true
        private: true
        token: <hf_token>
    
    graph_config:
      nodes:
        judge_pokemon:
          output_keys: pokemon
          node_type: llm
          prompt:
            - user:
                - type: text
                  text: |
                    Identify the pokemon in the provided image.
                - type: image_url
                  image_url: "{image}"
    
          model:
            name: gpt-4o
            parameters:
              max_tokens: 1000
              temperature: 0.3
      edges:
        - from: START
          to: judge_pokemon
        - from: judge_pokemon
          to: END
    
    output_config:
        output_map:
            id:
              from: "id"
            image:
              from: "image"
            pokemon:
              from: "pokemon"

\end{lstlisting}

\subsection{Pipeline to process audio inputs}
\begin{lstlisting}[style=yaml, frame=single]
    data_config:
      source:
        type: "hf"
        repo_id: "datasets-examples/doc-audio-1"
        split: "train"
        streaming: true
    
    graph_config:
      nodes:
        identify_animal:
          output_keys: animal
          node_type: llm
          prompt:
            - user:
                - type: text
                  text: |
                    Identify the animal in the provided audio.
                - type: audio_url
                  audio_url: "{audio}"
    
          model:
            name: qwen_2_audio_7b
            parameters:
              max_tokens: 1000
              temperature: 0.3
      edges:
        - from: START
          to: identify_animal
        - from: identify_animal
          to: END
    
    output_config:
        output_map:
            id:
              from: "id"
            audio:
              from: "audio"
            animal:
              from: "animal"

\end{lstlisting}

\end{document}